\documentclass[sigconf]{aamas}  
\pdfoutput=1

\usepackage{booktabs} 
\usepackage{amsmath}
\usepackage{graphicx}
\usepackage{algorithm}
\usepackage[noend]{algpseudocode}
\usepackage{amsfonts}
\usepackage{caption}
\usepackage{color}
\usepackage{balance}
\usepackage{subcaption}
\usepackage{placeins}
\usepackage{bbold}

\usepackage{todonotes}

\usepackage{array}
\newcolumntype{L}{>{\centering\arraybackslash}m{2.5cm}}

\begin{document}

\title{Safe Policy Search Using Gaussian Process Models}  




%
\author{Kyriakos Polymenakos}
\affiliation{%
  \institution{University of Oxford}
}
\email{kpol@robots.ox.ac.com}
\author{Alessandro Abate}
\affiliation{%
  \institution{University of Oxford}
}
\email{aabate@cs.ox.ac.uk}
\author{Stephen Roberts}
\affiliation{%
  \institution{University of Oxford}
}
\email{sjrob@robots.ox.ac.uk}

\renewcommand{\shortauthors}{K. Polymenakos et al.}

\begin{abstract}  

We propose a method to train a policy for the control of a dynamical system safely and with data-efficiently. We train a Gaussian process model to capture the system dynamics, based on the PILCO framework. The model has useful analytic properties, which allow closed form computation of error gradients and the probability of violating given state space constraints. Even during training, only policies that are deemed safe are implemented on the real system, minimising the risk of catastrophic failure.\footnote{This is an extended version of the paper with the same title presented in AAMAS 2019. See \url{http://www.ifaamas.org/Proceedings/aamas2019/pdfs/p1565.pdf}}

\end{abstract}

%


\maketitle


\section{Introduction}

Reinforcement learning (RL) is widely used to find suitable policies for a broad range of tasks. 
Model free methods are especially popular, but they often require a large number of interactions with the system,
usually a simulator, in order to converge to an effective policy. 
In applications without a reliable and accurate simulator, each policy has to be evaluated on a physical system. 
Physical systems limit the possible evaluations, since there is a non-negligible cost associated with each evaluation, 
in multiple resources: time, since every trial usually lasts several seconds, normal wear-and-tear inflicted to the system etc.

This scarcity of evaluations motivates us to be more data-efficient.
Model-based methods are known to often converge to a solution faster than model-free alternatives. 
However, models are also known to inhibit learning either by lack of flexibility or by introducing model bias, 
in both cases favouring solutions that are suboptimal, often persistently enough to stop the learning algorithm from finding better solutions. 
We address the lack of flexibility by using a non-parametric model which  can in principle be as flexible as needed. 
Model bias is also addressed, by using a model that explicitly accounts for uncertainty in its outputs. 
That way, even when the model's predictions are wrong, the model should provide them along with a suitably high uncertainty estimation, indicating that there is more to be learnt in the area and motivating further evaluations by the learning algorithm. 
Gaussian processes (GPs), our Bayesian non-parametric model of choice, 
fulfil both these criteria and are a suitable match for our purposes.
In fact, PILCO \cite{pilco}, is a framework for model-based, data-efficient reinforcement learning (RL), based on GPs. 
By meeting the above goals, it allowed for unprecedented data efficiency in a variety of tasks.

In many real application domains a primary concern is that of safety, particularly the avoidance of specific states, or sets of states, 
which are considered dangerous for the system or in general undesirable (for example, the avoidance of obstacles).
These constraints are defined \emph{a priori} and we require them to be respected, even during training. 
We make use of a Gaussian process model to estimate the risk of violating the constraints
before any candidate policy is implemented in the actual system.
 If the risk is too high we search for safer policies, implemented by tuning the optimisation process.
 Eventually, we obtain a set of the most promising of the policies that are further evaluated. 

\section{Relevant Work}

The task of choosing good parameter values for a controller is very general and widely encountered.
Control theory has been the go-to solution for designing controllers for dynamical systems, 
systems whose behaviour is described using a set of differential equations (or difference equations in the discrete case). 
In optimal control, usually a model is available and a cost function is defined. 
The objective is to minimise the expectation of the cost over a time horizon. 
Robust control deals with uncertain models, where some of the parameters are not known, but are usually bounded. 
In our case, we want to construct a model from scratch, 
from the data collected during training, allowing us to efficiently tune a controller. 

Reinforcement learning, \cite{sutton} is a machine learning framework which we may loosely see as lying between supervised and unsupervised learning. 
RL proposes that good policies are inferred by observing rewards (dependent on the actions taken) and adjusting the associated policy that maps states to actions and thence to future rewards.
There are model-based and model-free variants of RL approaches, with 
model-free methods usually being more flexible, whereas model-based methods offer more data efficiency.
Policies are described in two major ways in the majority of RL approaches.
Firstly, value function methods approximate a function that maps states (or state-action pairs) to expected returns. 
Secondly, policy gradient approaches use a parametrised policy and, by sampling trajectories from the system, 
estimate the gradient of a reward (or cost) function with respect to the policy parameters and propose a new value.
Methods that combine the above elements with some notion of safety, 
for example constraint satisfaction or variance reduction, form the subfield of safe reinforcement learning. 
For a review of safe RL see \cite{safe_survey}.

Our work expands the PILCO framework \cite{pilco,pil_thesis} a model-based approach to policy gradient reinforcement learning. 
The model of the system's dynamics that PILCO learns is based on Gaussian processes \cite{gpbook}. 
A distinctive trait of PILCO, in comparison to other policy gradient methods,
is that gradients are not numerically approximated from the sampled trajectories, 
but analytically calculated given the GP model. 
Interesting expansions have been made to the framework, 
including \cite{pilconn} which replaces the GP model with a neural network to lower the computational burden, 
with the most relevant to our work being that of \cite{pilco2}. 
Here, to discourage the system from visiting certain unwanted parts of the state space,
penalties are incorporated in the reward function, 
which successfully steer the system to valid regions of state space, without impeding learning. 
Other approaches in the literature use GP-based models to learn state space models,
as in \cite{gpdm} and \cite{identification}, 
but without observing perfect state information. 
This leads to a harder problem that calls for more complex architectures.

Another popular approach combines model-based learning methods with receding horizon control (usually Model Predictive Control, MPC).
GP-based models are used in \cite{model_rhc1, model_rhc2, pilcompc}, along with planning by solving
constrained optimisation problems online (during each episode), achieving good empirical performance. 
In \cite{kahn17}, neural networks with dropout are used instead of GPs to perform predictions \emph{with} uncertainty estimates. 

Furthermore, algorithms based on policy gradients have been employed for safe policy search, in \cite{trpo} and \cite{cpo}.
The policy updates are bounded based on the KL divergence (or approximations for it)
 between the old policy and the new one, and these bounds are used to encourage cautious exploration.
 Thus, catastrophic failure is avoided, as is shown both theoretically and experimentally. 
 These methods propose practical versions of the algorithms that scale well with more data and problems of higher dimensionality.

Bayesian optimisation has also been used to enforce safe policy search. 
In \cite{safe}, the authors suggest an algorithm that uses a surrogate model to avoid
 evaluating policies that have a high risk of performing under a set threshold. 
 The model is based on a GP, but does not capture they system's dynamics, 
 operating instead on the parameter space, as is usually the case in the Bayesian optimisation framework.
 In \cite{berk_quad} (and \cite{berk_swarm}) an extension is proposed,
 that allows the system to deal with more constraints (not only on the expected performance),
 as long as each evaluation of the policy returns a measure of how each constraint is to be violated. 
 Again, cautious exploration is encouraged, without using a simulator capable of predicting trajectories.
The probability of satisfying the safety property is calculated as an expectation over the posterior. 

An alternative approach to similar safety notions comes from the formal verification community.
For a rigorous treatment of systems' properties see for example \cite{model_checking}.
Recent works combine verification with learning from collected data \cite{eliz}.
In \cite{topcu}, safety properties are defined similarly but the authors assume a known model and unknown costs/rewards.
An SMT (Satisfiability Modulo Theory) solver is combined with RL to identify a set of safe policies,
and identify the optimal one belonging to the set.

The field of robotics is a natural domain of application for safety-aware learning methods. Here the common assumption is that a reasonably accurate model of the robotic system is available, but uncertainty can still be present, either in the motion of the system \cite{sun2016safe}, \cite{MC_planning}, or in the map, or model of the environment the robotic agent operates in  \cite{AxelrodRSS17}. 
Of particular interest to us is the work by \citet{akametalu2014reachability} which also uses a Gaussian process regression for safe learning. In their case, safety is achieved by the system switching when necessary to a safe control law, while in our case a single controller is combining both objectives. We also avoid solving the expensive Hamilton-Jacobi-Isaacs partial differential equation \cite{HJI} that is used to initially compute the safe control law.

Although much excellent work has been done in this domain, 
we regard previous work to lack some of the required components for our task. 
In particular,
in some cases a model is assumed to be known, albeit with incomplete knowledge of some parameters;
in other cases the problem is unconstrained, without concerns about safety. 
Finally, safety-focused methods place little emphasis on achieving good performance. 
Elsewhere, as in \cite{berk_quad}, the approach taken towards data-efficiency is distinctly different;
the system dynamics are not modelled, and a model is used to match the parameter space
(for example the values of the controller's gains) to the performance metrics directly.
This approach requires in general simple controllers with very few parameters, 
but can deal in principle with systems with complex dynamics.

\section{Problem Statement}
Our goal is to design a controller for an unknown, 
non-linear dynamical system that achieves good performance, 
as indicated by a reward function,
whilst avoiding unsafe states.

We assume:
\begin{itemize}
    \item a state space $X\subset \mathbb{R}^n$, 
    \item an input space $U \subset \mathbb{R}^m$ as the set of all legal inputs, 
    \item the dynamical system with a transition function $x_{t+1}= f(x_t,u_t) + v_N$,
    where $v_N$ is assumed to be i.i.d. Gaussian noise,
    \item a set $S \subset X$ of safe states and a corresponding $D = X \setminus S$ of unsafe (dangerous) states,
    \item a reward function $r : X  \rightarrow \mathbb{R}$.
\end{itemize}

Our task is to design a policy, $ \pi^\theta : X \rightarrow U $, with parameters $\theta$, 
that maximises the expected total reward over time T, 
while the system remains in safe parts of the state space at all times.
We require the probability of the system to lie in safe states to be higher than some threshold, $\epsilon > 0$.
The sequence of states the system passes through,
namely the \begin{it}trajectory\end{it}, is $ \mathbf{x} = \{x_1,...,x_T\} $ 
and we require all $x_i \in \mathbf{x}$ to be safe, meaning  that $x_i \in S$. 
Considering the probability associated with the trajectory as the joint probability
distribution of the $T$ random variables $x_i$, $p$, we focus on the probability:
\begin{equation}
\begin{split}
Q^\pi(\theta) &= \mathrm{Pr}(  x_1 \in S, x_2 \in S,..., x_T \in S ) \\
	 &= \int_S ... \int_S p(x_1, x_2,..., x_T)dx_1 dx_2 ... dx_T,
\end{split}
\end{equation}
which is the probability of all states in the trajectory being in the safe set of states $S$. We require that $ Q^\pi(\theta) > \epsilon$.

Our second goal is, as in all solutions, to maximise the performance of the system,
specifically by maximising the expected accumulated reward (return).
Once more this reward expectation is evaluated 
via the joint probability distribution over the set of the states the system passes through, namely:
\begin{equation} 
	R^\pi(\theta) =  \mathbb{E}_{X_T \sim p} \left [\sum_{t=1}^T r(x_t) \right ].
\end{equation}

\section{Algorithm}
\subsection{Model} \label{model}

We assume that the real dynamics of the physical system we want to control are given by
\begin{equation}
x_{t}= f(x_{t-1},u_{t-1}) + v_N,
\end{equation}
$$ v_N \sim \mathcal{N}(\mu,\Sigma).$$
Modelling the difference in consecutive states is preferable to modelling the states themselves 
for practical reasons, such as the fact that the common zero-mean prior of the GP is more intuitively natural.
It's calculated as $\Delta x_{t} = x_{t} - x_{t-1}$.

We chose the widely used squared exponential kernel for the covariance of the GP:
\begin{equation}
\begin{split}
 k(x_{1},x_{2}) &= \sigma_{s}^2 \text{exp}\left(-\frac{1}{2}(x_1-x_2)^T \mathbf{\Lambda}^{-1} (x_1-x_2)\right)\\
 		&+ \delta_{x_1,x_2} \sigma_{n}^2 .
 \end{split}
 \end{equation}
The squared exponential kernel choice reflects our expectation 
that the function we are modelling (the system dynamics) is smooth, 
with similar parts of the state space, along with similar inputs, to lead to similar next states. 
The kernel function's hyperparameters, namely the signal variance, length scales, and noise variance, 
are chosen through evidence maximisation \cite{gpbook}. 
Our approach uses off the shelf optimisers (BFGS or conjugate gradients), depending on numerical stability. 
Automatic Relevance Determination is employed and in cases with multiple output dimensions,
we train separate GPs on the data independently (but perform predictions jointly). 

The GP model, once trained, provides predictions for the mean and variance of $\Delta x_t$, given $x_t$ and $u_t$. 
To simplify notation, we denote the pairs of $(x_t, u_t)$ as $\bar{x_t}$, and so we have:

\begin{equation}
m_f(x_t) = \mathbb{E}[\Delta x_t] = k^T_\ast (K + \sigma^2_n I)^{-1}y = k^T_\ast \beta
\end{equation}

\begin{equation}
\sigma_f^2(x_t) = \text{var}_f[\Delta x_t] = k_{\ast\ast} - k_{\ast}^T( K + \sigma_n^2 I)k_{\ast}
\end{equation}

\noindent with $ k_\ast = k(\bar{X}, \bar{x}_t)$, $k_{\ast\ast} = k(\bar{x}_t, \bar{x}_t) $, and $\beta = (K + \sigma_n^2 I)^{-1}y$ where K is the Gram matrix of the kernel, with each entry $K_{ij} = k(\bar{x}_i, \bar{x}_j)$.

 To predict $x_{t+1},x_{t+2},\dots x_T$ we need to make predictions over multiple time steps. 
 To do that we have to use the output of the model ($\Delta x_t$) to estimate the current state $x_t$, 
 use the controller's parametrisation to estimate the control input $u_t$, 
 and feed the pair of state and input to the model as $\bar{x}_t$ to get the new prediction. 
 The next section explains this procedure in more detail.

\subsection{Multi-step prediction} \label{multi}

In this section we outline how the multi step predictions are made, following the analysis in \cite{pilco}. 

\sloppy Starting with an estimate of $x_{t-1}$, as a Gaussian distributed random variable
with mean $\mu_{t-1}$ and variance $\Sigma_{t-1}$ our goal is to estimate the state $x_{t}$. 
Firstly, using the policy $\pi^{\theta}$ we calculate the input $u_{t-1}$, as $\mu_u$ and $\Sigma_u$ (the policy is deterministic but the state is not). 
Then $\bar{x}_{t-1} = x_{t-1}$, $u_{t-1}$ is used to estimate $\Delta x_t$ using the GP model. We wish to estimate:
\begin{equation} \label{eq:intrac}
 p(\Delta x_t) = \int p(f(\bar{x}_{t-1}) | x_{t-1}, u_{t-1}) p(x_{t-1})dx_{t-1} ,
 \end{equation}
where we integrate out the variable $x_{t-1}$ and have dropped
 $p(u_{t-1} | x_{t-1})$, assuming a deterministic policy. 
This integral cannot be calculated analytically, neither it is tractable.
 Furthermore, an arbitrary distribution over $x_{t+1}$ would make the next prediction, $x_{t+2}$, even harder, etc. 
 To counter that, we use the moment matching approximation, where we calculate analytically the first two moments of 
$\Delta x_t$ and ignore all higher moments, approximating $p(\Delta x_t)$ with a Gaussian distribution. See more details in section \ref{noisy} of the Supplementary material.

Now let's assume that  we have a prediction for $\Delta x_t$, as $\mu_{\Delta x}$ and $\Sigma_{\Delta x}$. 
A prediction for $p(x_t)$ can be obtained via:
\begin{equation}
    \mu_t = \mu_{t-1} + \mu_{\Delta x}
\end{equation}
\begin{equation}
    \Sigma_{t} = \Sigma_{t-1} + \Sigma_{\Delta x} + \text{cov}[x_{t-1}, \Delta x_t] + \text{cov}[\Delta x_t, x_{t-1} ] \end{equation}
\begin{equation}     
    \text{cov}[x_{t-1}, \Delta x_t] = \text{cov}[x_{t-1}, u_{t-1}] \Sigma^{-1}_u \text{cov}[u_{t-1},\Delta x_t]
\end{equation}

The calculation of the covariances depends on the specific policy parametrisation, 
but for most interesting cases it's possible to perform them analytically.
For a more thorough explanation see  \cite{pil_thesis} and \cite{candela}. 
An illustrative example of the trajectory prediction step can be seen in Figure \ref{fig:trajectory}, where we plot three predicted trajectories for the collision avoidance experimental setup.

\subsection{Policy and risk evaluation}

So far, for any parameter value $\theta$, we can produce a sequence of mean and variance predictions
 for the states the system is going to be in the next T time steps. 
 We use this prediction to estimate the reward that would be accumulated by implementing the policy 
 and the probability of violating the state space constraints.
 We hence only evaluate promising policies, and we drastically increase the data efficiency of the method.

The sequence of predictions we get from the model (Gaussian posteriors over states) form a \textit{probabilistic trajectory}, which can be written as:
$$  tr^\theta = \{\mu_1, \Sigma_1, \mu_2, \Sigma_2, ..., \mu_T, \Sigma_T \} $$
Assuming a reward function of the form:

$$r(x) = \exp(- | x - x_{\text{target}} |^2 / \sigma_r),$$

since every predicted $p(x_t)$ is approximated by a Gaussian distribution we can write the total expected reward as:
\begin{equation}
    R^\pi(\theta) =  \mathbb{E}_{X_T \sim p} [\sum_{t=1}^T r(x_t)] = \sum_{t=1}^T\mathbb{E}_{x_t\sim p(x_t)}[r(x_t)] ,
\end{equation}
where:
\begin{equation}
    \mathbb{E}_{x_t}[r(x_t)]  = \int r(x_t) \mathcal{N}(x_t|\mu_t , \Sigma_t) dx_t.
\end{equation}

\noindent Similarly, for the probability of the system being in safe states during the episode, using the prediction for the trajectory:
\begin{equation} 
\begin{split}
Q^\pi(\theta) &= Pr(  x_1 \in S, x_2 \in S,..., x_T \in S ) = \\
&=\int_S ... \int_S p(x_1, x_2,..., x_T)dx_1dx_2...dx_T .
\end{split}
\end{equation} 
According to the moment matching approximation we are using, the distribution over states at each time step, is given by a Gaussian distribution, given the previous state distribution, so:
$$p(x_t | \mu_{t-1}, \Sigma_{t-1}) \approx \mathcal{N}(x_t | \mu_t, \Sigma_t) ,$$
and:nbb
\begin{equation}
\begin{split}
    Q^\pi(\theta) &\approx \int_S ... \int_S p(x_1) p(x_2 | x_1) ... p(x_T | x_{T-1}) dx_1 ... dx_T ,
\end{split}
\end{equation}
thus:
\begin{equation}
 Q^\pi(\theta) \approx \prod_{t=1}^T \int_S \mathcal{N}(x_t| \mu_t, \Sigma_t) dx_t = \prod_{t=1}^T q(x_t) ,
\end{equation}
where $q(x_t)$ is the probability of the system being in the safe parts of the state at time step $t$.
\begin{equation} \label{qs}
    q(x_t) = \int_S \mathcal{N}(x_t|\mu_t , \Sigma_t) dx .
\end{equation}

\subsection{Policy improvement}

After evaluating a candidate policy, our algorithm proposes an improved policy for evaluation.
This improvement can be a higher probability of respecting the constraints (making the policy safer) or an increase in the expected return.
As a secondary reward, we use a scaled version of the probability of respecting the constraints throughout the episode.
We investigated alternatives, and report the results in \Cref{sec:costFunctions}.

The objective function, capturing both safety and performance is  (a risk-sensitive criterion according to \cite{safe_survey}) is defined as : 
\begin{equation}
    J^\pi(\theta) = R^\pi(\theta) + \xi Q^\pi(\theta),
\end{equation} 
where we introduce hyperparameter $\xi$.

A gradient-based optimisation algorithm is used to propose a new policy.
The gradient of the objective function $J$ with respect to the parameters $\theta$ 
can be calculated analytically, using the model; 
the calculations are similar to the calculation of the reward gradient, 
which can be found in \cite{pilco} and in more detail in \cite{pil_thesis}.
The gradients are often estimated stochastically in the policy gradient literature (for example \cite{trpo}). 
However we do not have to resort to stochastic estimation, 
which is a major advantage of using (differentiable) models in general and GP models in particular.

The reward $R^{\pi}(\theta)$ accumulated over an episode, following PILCO, is a sum of the expected rewards received at each time step.
In order to calculate its gradient over the parameters the sum the gradients over all time steps.
The same applies when penalties are used to discourage visiting unsafe states, as in \cite{pilco2}.
In that case, instead of calculating a probability of being in an unsafe state,
the system receives an (additive) penalty for getting to unsafe states. 
The penalty is of the same form with the reward, and its gradients, gradients of the error with respect to the parameters, 
are calculated the same way, namely:
\begin{equation}
	\frac{d R^\pi(\theta)}{d \theta} =  \sum_{t=1}^T \frac {d \mathbb{E}_{x_t\sim p(x_t)}[r(x_t)]}{d \theta} .
\end{equation}
The probability of collision $Q^{\pi}(\theta)$, on the other hand, is a product of the probabilities of collision at every time step $q(x_t)$, hence:
\begin{equation}
	\frac{dQ^{\pi}(\theta)}{d\theta} = \sum_{t = 1}^{T} \frac{dq(x_t)}{d\theta} \prod_{j\neq t}^{N} q(x_j) .
\end{equation}
This formulation changes only the first step of the previous derivation. Here we only sketch the calculation of the gradients of the probability of collision at time step $t$, $q(x_t)$.

The partial derivative terms $ \frac{\partial q(x_t)}{\partial\mu(t)} $ and $\frac{\partial q(x_t))}{\partial \Sigma(t)}$ can be calculated fairly easily, assuming rectangular constraints. 

The terms $\frac{d\mu(t)}{d\theta}$ and $\frac{d\Sigma(t)}{d\theta}$ require us to consider how the parameters $\theta$ influence the mean and variance of the system's state at time $t$. To address we will have to use recursion, all the way to the starting state.
We here focus on the mean $\mu$ but the analysis for the variance is similar.
\begin{equation}
\begin{split}
\frac{d\mu(t)}{d\theta} &= 
    \frac{ \partial \mu(t)}{\partial\mu(t-1) } \frac{ d\mu(t-1) }  { d\theta }  \\
   &+ \frac{\partial \mu(t)} { \partial\Sigma(t-1) } \frac{ d\Sigma(t-1) }  { d\theta }  \\
   &+ \frac{\partial\mu(t)}{\partial\theta} .
\end{split}
\end{equation}
The terms $\frac{d\mu(t-1)}{d\theta}$ and $\frac{d\Sigma(t-1)}{d\theta} $ lead to a recursion as
they correspond to the required terms, but at the previous time step.
The partial derivative terms have to be calculated using the controller's and the GP's equations.
The gradients of the reward with respect to $\theta$ can be calculated in a very similar manner. 
Again, for more details we refer the reader to \cite{pilco} or \cite{pil_thesis}.

With the gradients of reward and risk in place, we can calculate the gradient of the full objective function $J$ and use it in a gradient based optimisation process.

\subsection{Safety check and adaptively tuning $\xi$}
When the optimisation stops, we have a new candidate policy to be implemented on the real system.
However before doing so, we want to make sure that it is safe.
It is possible for an unsafe policy to be optimal in terms of $J$, as long as the expected reward is high enough.
Here, we add a layer that in the event that the policy is unsafe,
disallows implementation,
increases $\xi$ by a multiplicative constant and restarts the optimisation's policy evaluation-policy improvement steps. 
Further, we check whether the policy is too conservative: 
if the policy is indeed safe enough we implement it,
and we also reduce $\xi$ by a multiplicative constant,
allowing for a more performance-focused optimisation in the next iteration of the algorithm. 

When the policy is implemented, we record new data from the real system. If the task is performed successfully the algorithm terminates. If not, we use the newly available data to update our model and repeat the process.  

This adaptive tuning of the hyperparameter $\xi$ guarantees that only safe policies
(according to the current GP model of the system dynamics) are implemented,
while mitigating the need for a good initial value of $\xi$. 
Indeed, using this scheme, we have observed that a good strategy is to start with a relatively high initial value, 
focusing on safety, which leads to safer policies that are allowed to interact with the system,
gathering more data and a more accurate model, and steadily discovering high performing policies as $\xi$ decreases.
For a succinct sketch see \Cref{al:1}.

\begin{algorithm}
    \caption{Main SafePen algorithm}\label{al:1}
    \begin{algorithmic}[1]

        \State $\text{Initialize } \theta, \xi$
        \State $\text{Interact with the system, collect data}$
        \State $\text{Train GP model on the data}$
    
        \Repeat
        \Repeat
        \State $\text{Evaluate policy as } J^{\pi}(\theta) = R + \xi Q$
        \State $\text{Update policy using gradient of $J^{\pi}(\theta)$}$
        \Until $\text{Convergence or a time limit is reached}$
    
        \State $\text{Calculate } Q$
        \If {$Q > \epsilon$}
        \If {$Q > \text{upper\_limit} $ }
        \State Decrease $\xi$
        \EndIf
        \State $\text{Interact with the system, collect data}$
        \State $\text{Retrain GP model on the new data set}$
        \Else
        \State $\text{Increase } \xi$
        \EndIf
        \Until task learned (or run out of time, interactions budget etc.)
    \end{algorithmic}
\end{algorithm}

\section{Experiments}
\subsection{Simple Collision Avoidance}\label{sec:mainExp}
In this scenario two cars are approaching a junction. 
We are assuming the point of view of one of the two drivers with the objective of crossing the junction safely by accelerating or braking. 

The system's state space $ \mathcal{X} $ is 4 dimensional (2 position and 2 velocity variables): 
$$ x_t = [x_{1t}, ... , x_{4t} ]. $$

The input $u$ has one dimension, proportional to the force applied to the first car.

In this application the differentiation between safe and unsafe states is intuitive and straightforward. In order to not collide, the cars must not be at the junction (set to be the origin (0,0)) at any point in time. This can be set as a constraint of the form: 
$$ |x_{1t} | > a \quad  \mathbf{OR}  \quad | x_{3t} | > a $$
for the two cars positions.  We hence denote the set of safe states as,
$$ S = \{x \in X : \ |x_1| > a \  \mathbf{OR} \ |x_3| > a  \},$$
where $a$ is a reasonably valued constant (10m for example).

If we ignore velocities and the think of the state space as a plane with the cars' positions on the axis, the unsafe set of states forms a rectangle around the origin.
The legal inputs are one dimensional corresponding to the force applied to the controlled car (car1) which we assume bounded at 2000N,
accelerating or decelerating the car.

The controller used is a normalised RBF network with 20 units as in \cite{pilco}, initialised randomly
and the reward function is an exponential centred on after the junction, 
parametrised only on the position of car1.

The first approach we employ is inspired by \cite{pilco2}, and is a variant of the standard PILCO framework,
adding penalties on states that need to be avoided. 
The penalties are smooth, based on exponential functions, much like the rewards.
The authors show that a robotic arm is able to stack building blocks constructing a small tower,
without destroying the tower when trying to add blocks. 
\begin{figure*}[h]
\begin{minipage}[t]{0.5\textwidth}    
	\begin{figure}[H]
    		\includegraphics[width=\textwidth]{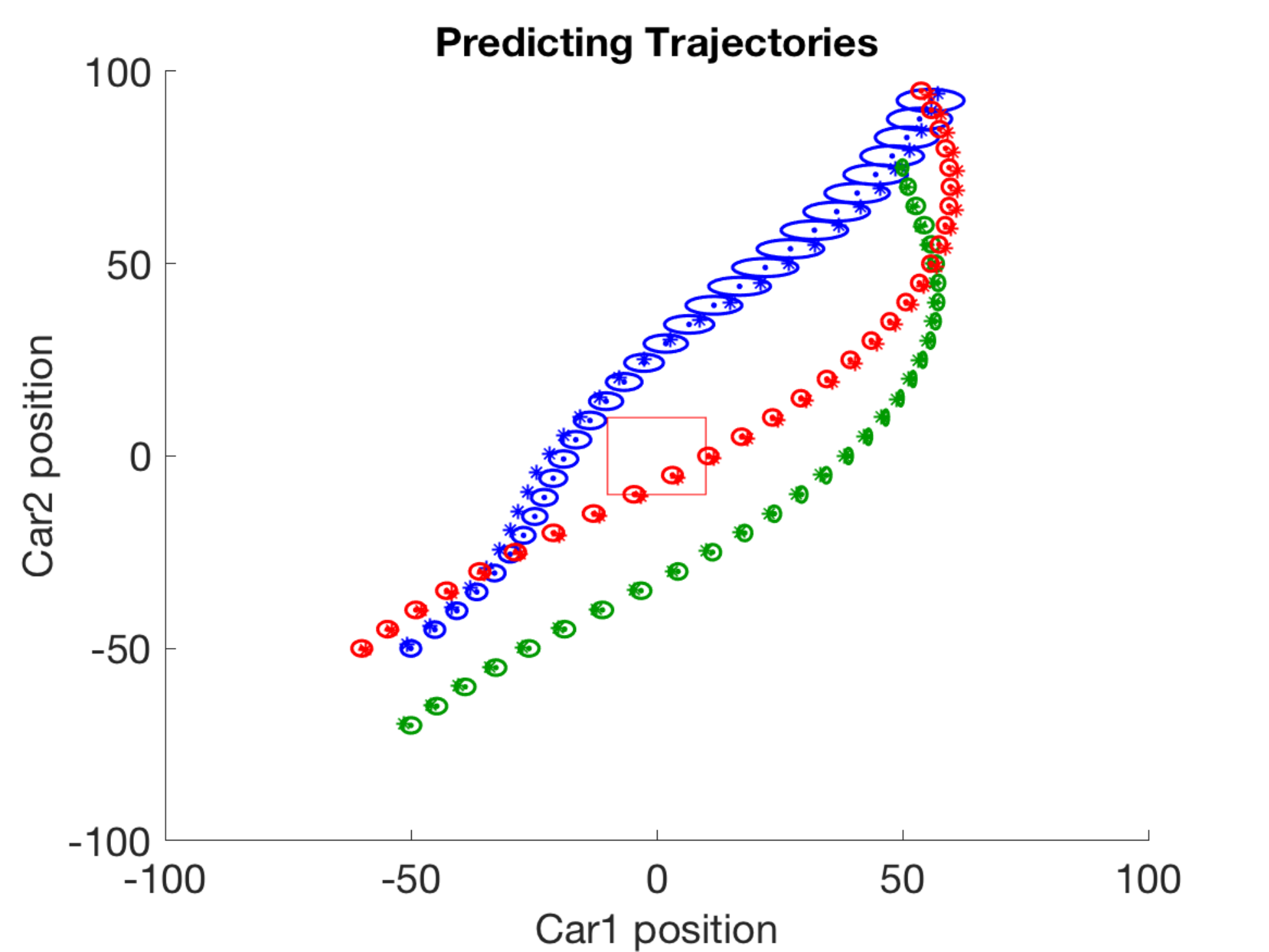}
    		\centering
    		\captionsetup{width=0.9\textwidth}
    		\caption{\small{Three trajectories and associated predictions. In blue an inaccurate model captures the underlying uncertainty in a scenario where the first car accelerates and passes through the junction first. In green, an accurate model predicts well the trajectory in a scenario where the second car crosses the junction first. In red, both cars cross the junction at the same time, resulting in a collision. } }
	\label{fig:trajectory}
	\end{figure}
\end{minipage}%
\begin{minipage}[t]{0.5\textwidth}
	\begin{figure}[H]
    		\includegraphics[width=\textwidth]{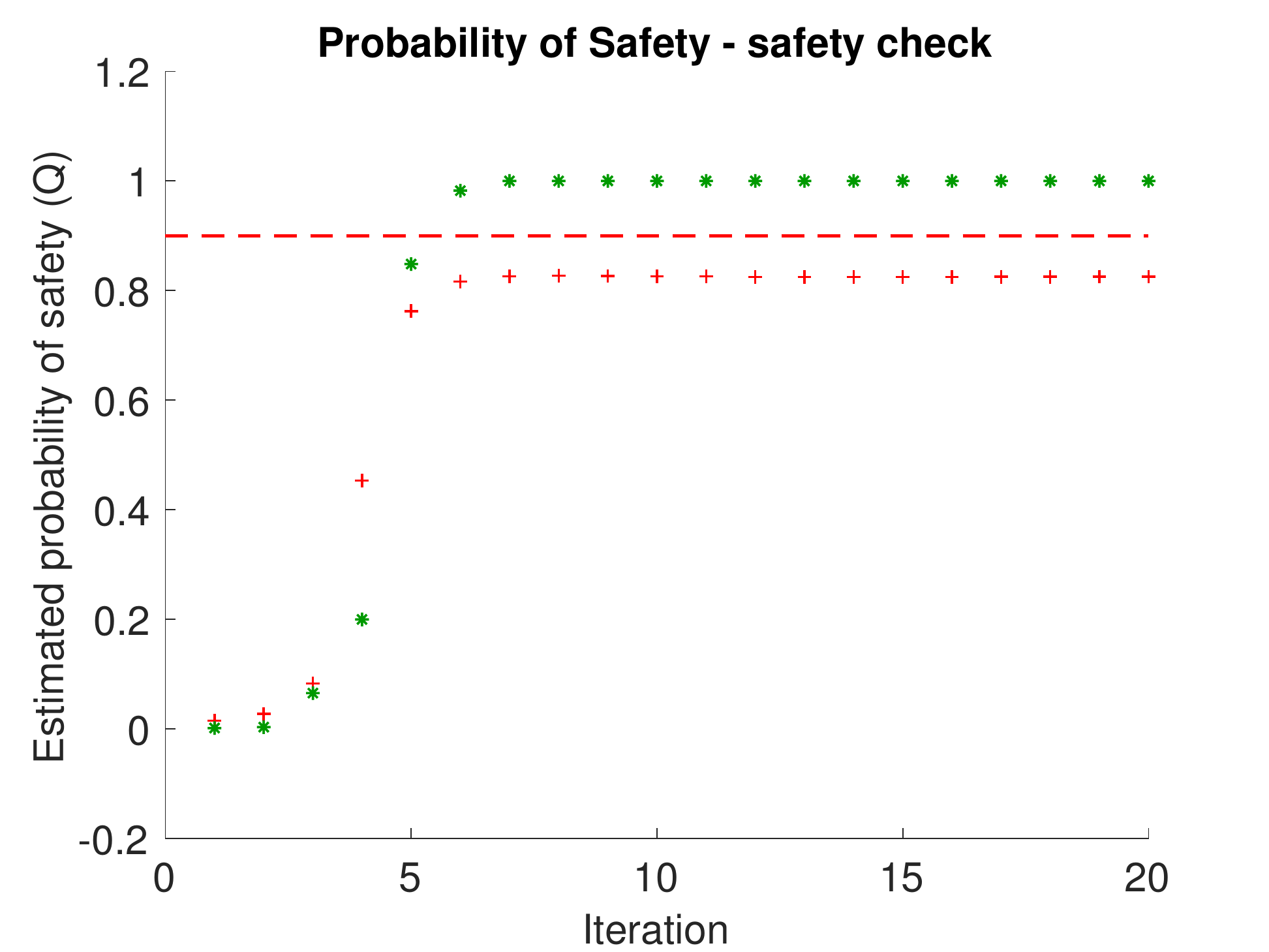}
    		\centering
    		\captionsetup{width=0.9\textwidth}
    		\caption{\small{Evaluations of $Q^\pi(\theta)$, at step 9 of Algorithm 1, for two training runs. In the first case, in red, $Q$ is less than the threshold $\epsilon = 0.9$ for the the first 5 policies proposed. Onwards, safe enough policies are proposed by the algorithm. With green we highlight a potentially problematic scenario, where the algorithm fails to propose a safe enough policy after 20 cycles. Still collisions are avoided, since while $Q^\pi(\theta) < \epsilon$ interaction with the physical system is prohibited. Experimentally, this behaviour is uncommon but possible.}}
	\label{fig:safety}
	\end{figure}
\end{minipage}
\end{figure*}

In \Cref{table:main} we see a performance comparison for our proposed method (SafePen) and the baseline PILCOPen. 
For PILCOPen, we report results for different values of the hyperparameter $\xi$. 
It is obvious how sensitive the baseline is to the choice of $\xi$. 
The proposed method, by not allowing unsafe policies to be implemented, while optimising the $\xi$ value adaptively, 
achieves good performance, with a relatively low number of collisions, and with the hyperparameter initialisation not having a significant effect. Furthermore, SafePen interacts with the system fewer times, due to avoiding interactions when the policy is not safe enough. 
In one case out of the 48 runs, no interaction took place, since the algorithm failed to propose a safe policy (see \Cref{fig:safety}).

\begin{table}
\centering
\resizebox{0.75\linewidth}{!}{%
\begin{tabular}{c c c c c c}
\hline
 & SafePen & \multicolumn{3}{c}{ PILCOPen} \\
$\xi$ & - & 1 & 10 & 20 \\
collisions & 30 & 614 & 32 & 81   \\ 
av. cost & 8.89& 5.06 & 9.87 & 9.96   \\  
interactions & 526 & 720  & 720 & 720\\ \hline \\
\end{tabular}}
\captionsetup{width=0.45\textwidth}
\caption{\small{Comparison between SafePen (our algorithm) and PILCOPen. The number collisions captures overall safety during training, while the average cost refers only to the performance component $R$ of the objective function $J$ (a convention we follow for the rest of the paper). Both methods are evaluated on 48 runs, with a maximum of 15 interactions with the real system per run.}}
\label{table:main}
\end{table}

 \subsubsection{Surrogate loss functions for the safety objective}\label{sec:costFunctions}

As discussed previously, the probability of the system violating the constraints during an episode has a dual role:
 \begin{itemize}
 \item In the policy evaluation - policy improvement iterations, it is a component of the objective function, along with the expected reward.
 \item In the safety check step, where it is evaluated in order to decide whether the policy is safe enough to be  implemented in the physical system.
 \end{itemize}

In the problem formulation we require keeping the probability of violating the constraints under a tolerable risk threshold 
every time the learning algorithm interacts with the physical system. 
This requirement corresponds to the second case described above.
In the first case, on the other hand, evaluating this probability is not a necessity so long as the policy resulting from the optimisation procedure verifies this safety requirement. 
We therefore explored substituting this probability, namely $Q^\pi(\theta)$, with a surrogate loss function which we
denote $\hat{Q}^\pi(\theta)$. We investigate whether this substitution increases computational efficiency without impeding performance. 
Increased performance is expected if the surrogate creates a loss landscape that facilitates the optimisation process. 
Indeed, we can interpret the exponential penalties used in \cite{pilco2} as an example of a surrogate loss function 
and we evaluate its performance in the experiments that follow.

Two loss functions have been considered thus far: a scaled version of the probability itself (no surrogate in this case) and  exponential penalties. For completeness, we consider two more: an additive cost function based on the sum of the probabilities of violating the constraints on each time step (ProbAdd for short) 
and one based on a logarithmic transform of the original (multiplicative) probability (LogProb). 

The additive cost based on the risk of constraint violation (ProbAdd) is defined as: 
\begin{equation}
  \hat{Q}^\pi(\theta)_{\text{add}}  = \sum_{t=1}^T q(x_t) ,
\end{equation}
with $q(x_t)$ as defined in (\ref{qs}).

Assuming perfect safety is feasible, with all $q(x_t) = 1$, 
maximising this surrogate loss function leads to a maximisation of the original objective, which is the product of  $q(x_t)$ for all $t$. 
When this is not feasible on the other hand, the maximum is not necessarily the same. 
Keeping in mind that convergence on the global maximum is not guaranteed for any of the methods used here for policy optimisation,
the effectiveness of the surrogate cost function is evaluated empirically.

Taking the logarithm of the original multiplicative probability allows us to create an additive cost function,
while maintaining the same maximum, since the logarithm is a concave function, defined as:
\begin{equation}
  \hat{Q}^\pi(\theta)_{\text{log}}  = \log \prod_{t=1}^T q(x_t) = \sum_{t=1}^T \log (q(x_t)) .
\end{equation}
Differentiating the above is not significantly different than the process described previously and the gradients are used in the same optimisation algorithm.
 
We present results  in \Cref{table:costFunctions}. The four cost functions are evaluated on 32 runs,
with 15 maximum allowed interactions per run with the physical system (480 interactions allowed in total).
We can see that all methods achieve a fairly low number of collisions (much lower than the maximum allowed risk of 10\% per interaction), 
and the Prob and Log surrogate cost functions achieve slightly better performance. 
The Log and the ProbAdd cost functions fail to propose a controller for a non-negligible number of cases (4 and 6 out of 32), 
a fact suggesting that the optimisation proved harder using these cost functions. 
For the rest of this paper the Prob cost function is used (and thus denoted simply SafePen, except when explicitly stated otherwise), since it achieves good performance and is in general a simpler approach, 
in the sense that objective (low risk) and cost function match, other than the multiplicative constant ($\xi$).

\begin{table}
\centering
\resizebox{0.8\linewidth}{!}{
\begin{tabular}{c c c c c c}
\hline
 & Prob (\textbf{SafePen}) & Exp & Log & ProbAdd \\
collisions & 12 & 8 & 4 & 11   \\ 
av. cost & 8.75 & 9.19 & 8.78 & 9.28    \\  
interactions & 327 & 412 & 166 & 315\\ 
unsolved & 0 & 0 & 4 & 6 \\ \hline \\
\end{tabular}}
\captionsetup{width=0.45\textwidth}
\caption{\small{Surrogate cost function comparison. `Prob' denotes the probability of collision used as a multiplicative cost, `Exp' smooth exponential penalties, `Log'  the log of the probability of collision, and `ProbAdd'  the sum of the probabilities of collision at each time step (additive cost).}}
\label{table:costFunctions}
\end{table}

\subsubsection{Hyperparameter tuning}
Here we examine the effects of the method we introduce for tuning the hyperparameter $\xi$. 
Using the same objective function, we change only the way $\xi$ is set and tuned, 
isolating and evaluating its effects on performance.
The simplest version we evaluate uses a fixed value for $\xi$, as in \cite{pilco2} and the PILCOPen algorithm we employed in \Cref{sec:mainExp}. 
We compare the latter with a version of our algorithm that  checks whether the policy is safe enough,  increasing $\xi$ 
if not (but never decreasing it).
Finally, we compare with a version of the algorithm
that adapts $\xi$  by increasing or decreasing its value accordingly (SafePen in \Cref{sec:mainExp}). 
To provide a fairer comparison we use the same surrogate loss in all cases, 
namely that of exponential penalties on the unsafe parts of state space.

We see in \Cref{table:tuning} that an adaptively tuned $\xi$ leads to similar, or better, performance and more robustness to the initial choice of $\xi$.
\begin{table}
\centering
\resizebox{0.8\linewidth}{!}{%
\begin{tabular}{c c c c c c} \hline
 & \multicolumn{2}{c}{ PILCOPen} & PenCheck & \multicolumn{2}{c}{ SafePen} \\
(starting) $\xi$  & 5 & 10 & 5 & 5 & 10  \\
collisions & 36 & 0 & 0 & 1  & 2  \\ 
av. cost & 7.54 & 13.31 & 14.45 & 14.51 & 12.66   \\ 
interactions & 240 & 240 & 138 & 97 & 138 \\ \hline \\
\end{tabular}}
\captionsetup{width=0.45\textwidth}
\caption{\small{Different strategies of setting $\xi$ and their effects for different initial values. PILCOPen picks one value for $\xi$ and implements all policies that are proposed by the policy optimisation algorithm, PenCheck uses the safety check before implementing a policy and increases $\xi$ if the policy is unsafe, and SafePen, increases and decreases the hyperparameter adaptively.}}
\label{table:tuning}
\end{table}

\subsection{Building Automation Systems} \label{sec:BASB}

We here apply our approach to a problem in the domain of \emph{building automation systems},
often referred to as smart buildings.
The usual aim here is to efficiently use the air conditioning system,
keeping the temperature and possibly $\mathrm{CO}_2$ emission levels within given limits,
or close to specified values. 
These values often correspond to comfortable conditions for occupants,
but can vary significantly depending on the building's purpose.
As the number of available sensors and their connectivity increase, so does the
sophistication of the controlling mechanisms of these systems.
Increased emphasis on energy consumption, with its associated financial and environmental cost, 
and of course the desire to improve the experience of those in the buildings, contribute to the rising
interest in the area \cite{cauchi2018benchmarks}. 
Finally, it's worth noting that one of the most well-known real-world applications of reinforcement learning was in
this domain, with Deepmind offering an RL optimised system for cooling Google's data centres \cite{deepmindBlog}.

For our experiment we use as ground truth an open source simulator 
\footnote{Code for the simulator can be found at \url{https://gitlab.com/natchi92/BASBenchmarks}.} 
that models air conditions, 
released as part of a set of benchmarks \cite{cauchi2018benchmarks, abate2018arch} for the verification of stochastic systems.
Following the same approach outlined in section \ref{sec:mainExp}, we treat the simulator as a black box: 
our algorithm sees the data generated, as sequences of states and actions, 
but has no access to the simulator's internal parameters.
With the data collected, we train a GP-based dynamics model, and follow Algorithm \ref{al:1}.

In this setting, matching Case Study 2 from \citet{cauchi2018benchmarks}, we control the temperature in two adjacent rooms. 
The cost we minimise is the quadratic error between the temperatures in the two rooms and the reference temperature. 
The total number of state variables is 7 (including room temperatures, wall temperatures etc.)
while the control input we have at our disposal is one-dimensional 
and corresponds to the (common) air supply temperature for the two rooms. 
The measurements have a sampling period of 15 minutes. 
We collect 72 hours worth of data to start training, and use a simple linear controller. 
The initial temperature in room 1 is $15^\circ$C and the target temperature is $20^\circ$C, while room 2 starts with the target temperature.
The task is to gradually increase temperature in room 1, while keeping the temperature in room 2 below $20.5^\circ$C. 
Since the air supply for the two rooms is shared, aggressive temperature control for room 1 would lead to the temperature in room 2 overshooting both the target of $20^\circ$C, and the constraint of $20.5^\circ$C, a hypothesis we verify by running 
plain PILCO.

\begin{figure}
	\centering
	\includegraphics[width=.95\linewidth]{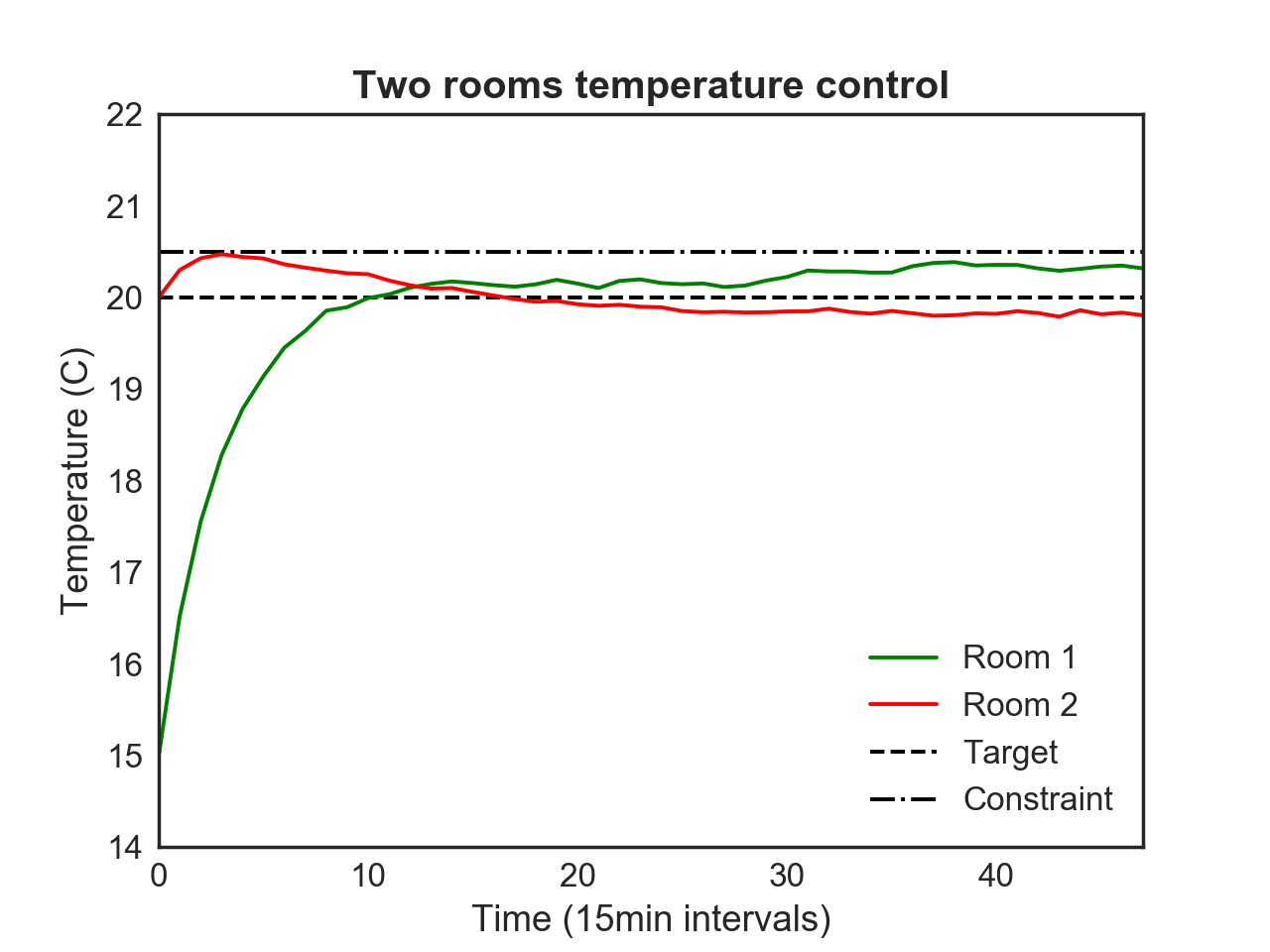}
	\caption{\small Temperature control scenario where room 1 starts with a significantly lower temperature than the target.
					The linear controller used has been trained with our algorithm (SafePen).}
	\label{fig:basb2}
	\vspace{-10pt}
\end{figure}

In Table \ref{table:BASB1}, we summarise the results obtained by plain PILCO, optimising performance exclusively, 
PILCOPen, using an exponential penalty of preset weight to encourage safety, 
and SafePen, using the adaptively weighted penalty and the safety check we introduced.
All algorithms interact with the system for 5 episodes of 12 hours each. The results are averaged over 10 random seeds.
We can see in \ref{table:BASB1} that SafePen manages to respect the constraints without incurring practically any extra cost,
while PILCOPen (using a hand-crafted reward, whose shape and weight were tuned to good values to the best of our knowledge)
respects constraints but with significant additional cost. 
Plain PILCO is included to show the performance of unconstrained policies, in terms of both cost and constraint violations.
In Figure \ref{fig:basb2} we see a typical response of the two-room system (as predicted by the simulator) to the controller 
trained with the proposed algorithm (SafePen).

\begin{table}
	\centering
	\resizebox{\linewidth}{!}{
		\begin{tabular}{c c c c} \hline
			& PILCO & SafePen (Add) & PILCOPen ($\xi=2$ )\\
			Con. Viol. (steps) & 26.1 $\pm$ 8.7 & 0.0 $\pm$ 0.0 & 0.0 $\pm$ 0.0\\ 
			Con. Viol. (epis.)  & 4.5   $\pm$ 1.5 & 0.0 $\pm$ 0.0 & 0.0 $\pm$ 0.0\\  
			RMSE 			   & 0.94 $\pm$ 0.32 & 0.95 $\pm$ 0.31& 1.24 $\pm$ 0.41\\ 
			\hline \\
	\end{tabular}}
	\captionsetup{width=0.45\textwidth}
	\caption{\small{Results for the BAS environment.
				Regarding constraint violations, we report the number of time steps where 
				constraints were violated during the 5 episodes of interaction in Con. Viol (steps), 
				as well the number of unique episodes with at least one constraint violation in Con. Viol (epis.).
				The RMSE refers to the temperature difference of the two rooms from the target value and is used as an interpretable cost.
				Results are averaged over 10 random seeds, with one standard deviation shown. 
					}}
	\label{table:BASB1}
\end{table}

\subsection{OpenAi Gym experiments - Swimmer}

\begin{figure}
	\begin{subfigure}{.23\textwidth}
		\centering
		\includegraphics[width=.8\linewidth, height=3cm]{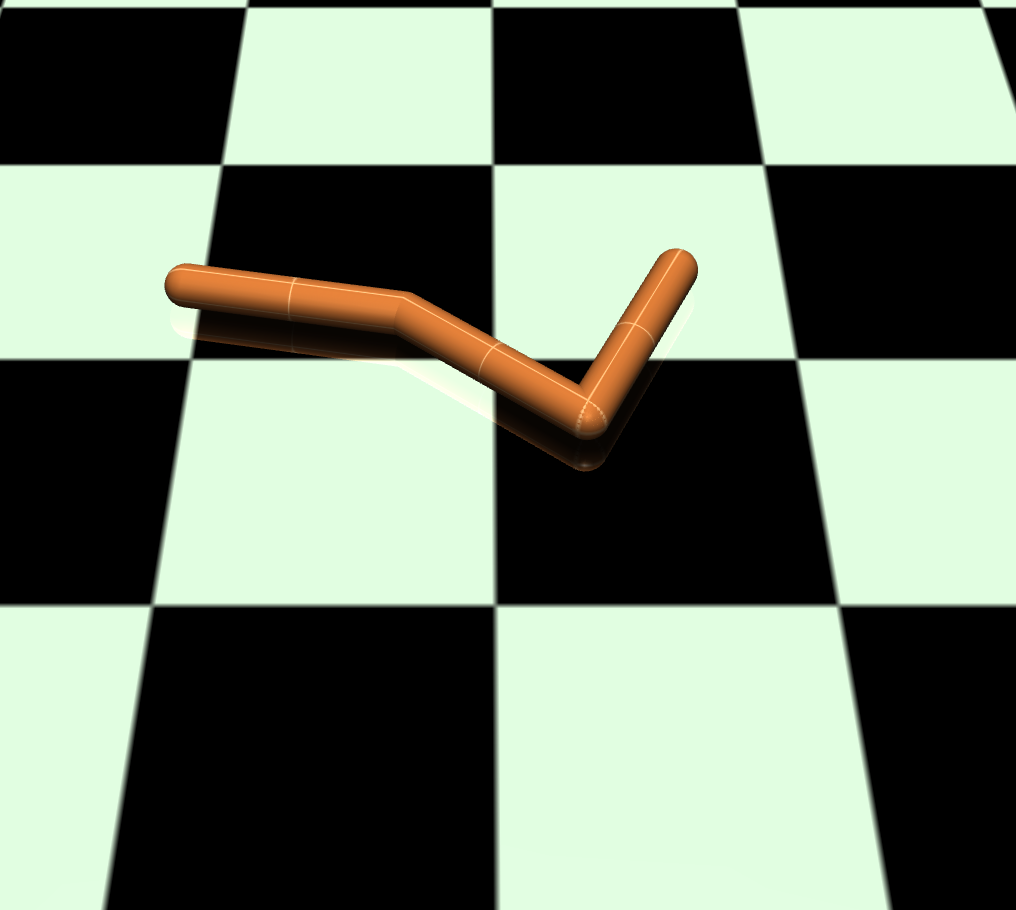}
		\caption{\small Max angle at the first joint}
		\label{fig:sfig2}
	\end{subfigure}%
	\begin{subfigure}{.23\textwidth}
		\centering
		\includegraphics[width=.8\linewidth, height=3cm]{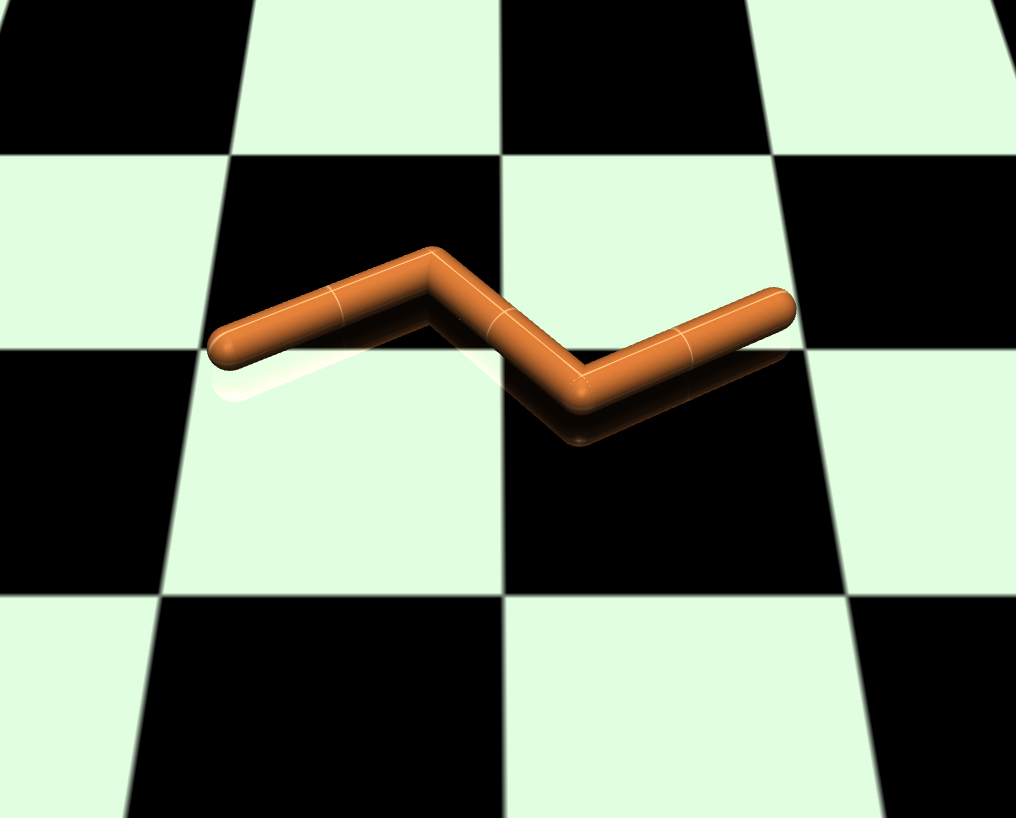}
		\caption{\small Frame from a healthy gait}
		\label{fig:sfig3}	
	\end{subfigure}
	\caption{\small Various configurations of the Swimmer Robot}
	\label{fig:Swimmer}
\end{figure}

Next we evaluate our method using a challenging task 
from the popular OpenAi-gym set of benchmarks for RL. 
Specifically we work with Swimmer-v2, a scenario where a robot with two joints
navigates a 2-d plane by "swimming" in a viscous fluid. 
Both joints are controlled by an actuator, and the task requires the robot to 
move forward on the x-axis. 
The OpenAi state space is 8-dimensional, with the state variables corresponding to:
orientation of the first link, angle of the first joint, angle of the second joint, 
velocity of the end point on the x-axis, velocity of the end point on y-axis and finally the angular velocities for the first three state variables. 
The control is 2-d, and corresponds to torque applied by the two actuators associated with the two joints.
This scenario represents a significantly harder challenge for the learning algorithm since it is higher dimensional, non-linear, 
and under actuated, making it harder to acquire a strong reward signal at the early stages of training \cite{guided_journal}.

The scenario does not natively provide safety constraints. 
We therefore impose constraints on the mechanism by limiting the angles of the two joints to a maximum value, with the intuition being that 
pushing the joints to the edge of their working range can lead to damage,
either from the accumulated stress to the joints themselves or by having parts of the robot collide. 
Furthermore, an interesting qualitative observation is that in many runs of the simulation,
in the absence of constraints,
a gait emerges that has the robot bend its first joint with full force,
effectively using it as single paddle, 
gaining significant speed and getting high reward early in the episode,
but getting stuck in the resulting configuration - 
often with full force still being applied to one or both joints (see \ref{fig:sfig2}).
We consider this a characteristic example of the pitfalls that come with the use of RL for 
where failure can have significant costs.

We evaluate two variants of PILCOPen and our proposed algorithm SafePen \footnote{We are using the Python and Tensorflow based implementation that 
can be found at \url{https://github.com/nrontsis/PILCO}}.
In these experiments, every episode has 25 time steps (corresponding to 125 steps in the original environment since we subsample by a factor of 5), 
all algorithms obtain data from $J=10$ episodes from a random policy at first and then a maximum of $N=10$ interactions with the environment. 
The best return metric corresponds to the highest return of these $N$ interactions, averaged
over 10 runs with different random seeds.
Similarly the number of constraint violations corresponds to 
the number of time steps with a constraint violation during training 
(for the $N$ episodes where a trained policy is implemented and not the initial random rollouts),
averaged over 10 random seeds.
We also report the number of episodes where at least one constraint violation occurred, out of the
10 interactions with a simulator the algorithms are allowed. 

Table \ref{table:swimmer1} summarises the experimental results. The proposed algorithm SafePen, ends up violating the constraints with the maximum allowed 
frequency ($\epsilon = 0.1$ here, and we have 1 constraint violating episode out
of 10 on average). PILCOPen, when using a small penalty, achieves better performance
with a higher number of constraint violations, while a higher penalty leads to safer 
behaviour and decreased performance, as expected.

\begin{table}
	\centering
	\resizebox{\linewidth}{!}{%
		\begin{tabular}{c c c c c} \hline
			& \multicolumn{2}{c}{PILCOPen} & SafePen (Add) \\
			$\xi$  & 1 & 10 & -  \\
			Con. Violations (steps) & 59.2 $\pm$ 28.8 & 1.3 $\pm$ 3.6 & 3.2 $\pm$ 4.07 \\
			Con. Violations (epis.) & 7.4 $\pm$ 2.6  & 0.3 $\pm$ 0.64 & 1.0 $\pm$ 1.0 \\  
			Best Return & 10.63 $\pm$ 0.87  & 4.80 $\pm$ 2.09 & 7.63 $\pm$ 2.15 \\ 
			\hline \\
	\end{tabular}}
	\captionsetup{width=0.45\textwidth}
	\caption{\small{Results for the swimmer environment. 
					Con. violations (steps) refer to the total number of time steps where a 
					constraint is violated during training, 
					while con. violations (epis.) counts the different episodes (out of 10) where 
					a constraint violation occurred. 
					All results are averaged over 10 runs, and reported along with one standard deviation.
					}}
	\label{table:swimmer1}
	\vspace{-10pt}
\end{table}
 
\section{Discussion and Future Work}

In this work we propose a method to integrate model-based policy search with safety considerations throughout the training procedure.
Emphasis is given to data-efficiency, since we require our approach to be suitable for applications on physical systems.
Using a state-of-the-art PILCO framework, we incorporated constraints in the training,
estimating the risk of a policy violating the constraints and preventing high risk policies from being applied to the system.
Our  contribution uses probabilistic trajectory predictions obtained as model outputs in two ways:
(a) to evaluate the probability of constraint violation and 
(b) as part of a cost function, to be combined with performance considerations.
Furthermore, the proposed adaptive scheme successfully allows a trade-off between the two objectives of safety and performance, 
alleviating the need for extensive hyperparameter tuning.
An important future challenge would be applying our method to real-world systems, 
with all the additional complexity that comes with moving away from simulation.
Nevertheless, the combination of data efficiency and safety awareness renders the method well-suited
for physical systems.
FInally, investigating alternatives to the current modelling approach, 
namely a global, RBF-kernel based GP model for the system dynamics, is a promising research direction.

\section*{Acknowledgements}
This work was supported by the EPSRC AIMS CDT grant EP/L015987/1 and Schlumberger.

\FloatBarrier

\bibliographystyle{ACM-Reference-Format}  
\balance
\bibliography{cite}  

\section{Supplementary material}

\subsection{Gaussian Process regression on noisy inputs} \label{noisy}

As we saw in \ref{model}, given an input $x_\ast$ the GP can give as a predictive distribution, in the form of a Gaussian, for the output, which in our case is the predicted difference between $\Delta x$. Since predictions have to be cascaded, as mentioned in \ref{multi}, instead of a point $x_\ast$ we have Gaussian distributed input, $x_{t-1} \sim \mathcal{N}(\mu_{t-1}, \Sigma_{t-1})$.
That gives rise to the intractable integral in Equation \ref{eq:intrac}, which we approximate with a new Gaussian distribution, using moment matching, by (analytically) computing the mean and variance of the predictive distribution \cite{candela}. We include this analysis here, following \cite{pilco}, for the sake of completeness. 
Assuming $d = 1,\dots,D$ target dimensions for the mean of the predictive distribution, we have:

\begin{equation}
\mu_\Delta^d = \mathbb{E}_{x_{t-1}} [ \mathbb{E}_f [f(x_{t-1}|x_{t-1}) ]] = \mathbb{E}_{x_{t-1}} [m_f(x_{t-1})] 
\end{equation} 
which has been to shown to be:
\begin{equation}
\mu_\Delta^d = \beta_d^Tq_d
\end{equation}
with $\beta_d = (K_d + \sigma_{n_d}^2)$, and $w_d = [w_{d1}, \dots, w_{dn}]^T$ where each $w_{di}$ is:
\begin{equation}
w_{di} = \frac{\epsilon_d^2}{ \sqrt{ | \Sigma_{t-1}\Lambda_d^{-1} + I} | } \exp\left( - \frac{1}{2} v_i^T ( \Sigma_{t-1} + \Lambda_d)^{-1}v_i\right)
\end{equation}
and:
\begin{equation}
v_i = (x_i - \mu_{t-1})
\end{equation}
We now want to calculate the predicted covariances for the D targets, each covariance matrix being $D \times D$. 
It is here that the kernels learnt for the different dimensions are combined to produce predictions, 
otherwise we would be able to treat the GP model as $D$ independent models, multi-input, single-output GP regressors.

For diagonal elements we have:
\begin{equation}
	\sigma_{dd}^2 =  \mathbb{E}_{x_{t-1}}[\text{var}_f[\Delta_d | x_{t-1}]] + \mathbb{E}_{f,x_{t-1}}[\Delta_d^2] - (\mu_\Delta^d)^2
\end{equation},
while for non-diagonal elements:
\begin{equation}
	\sigma_{d_1d_2}^2 = \mathbb{E}_{f, x_{t-1}} [\Delta_{d_1} \Delta_{d_2}] - \mu_\Delta^{d_1} \mu_\Delta^{d_2}
\end{equation}
where $d_1$ and $d_2$ are two different target dimensions.

Comparing the two expressions we can see that the term \\
$ \mathbb{E}_{x_{t-1}}[\text{cov}_f[\Delta_{d_1}, \Delta_{d_2} | x_{t-1}]] $ is missing, 
and this exactly because the two predictions $\Delta_{d_1}$,  $\Delta_{d_2}$ are independent, \emph{when} $x_{t-1}$ \emph{is given}. 
For the rest of the terms, we have:
\begin{equation}
\mathbb{E}_{f, x_{t-1}} [\Delta_{d_1} \Delta_{d_2}] = \beta_d^T W^{d_1 d_2} \beta_d
\end{equation}
with $W^{d_1 d_2} \in \mathbb{R}^{N \times N}$, with $N$ the dimensionality of the GP input,  and each entry being:
\begin{equation}
W_{ij}^{d_1 d_2} = \frac{k_{d_1}( x_i, \mu_{t-1})  k_{d_2}(x_j, \mu_{t-1}) }{\sqrt{|R|}} \exp( \frac{1}{2} z_{ij}^T R^{-1} \Sigma_{t-1} z_{ij} ) 
\end{equation}
where:
\begin{equation}
R = \Sigma_{t-1} (\Lambda_{d_1}^{-1} +\Lambda_{d_2}^{-1}) + I
\end{equation}

\begin{equation}
z_{ij} = \Lambda_{d_1}^{-1} v_i + \Lambda_{d_2}^{-1} v_j
\end{equation}
\noindent The $\beta_d$ and $v_i$ are the same used for the mean value calculation. Finally:

\begin{equation}
 \mathbb{E}_{x_{t-1}}[\text{var}_f[\Delta_d | x_{t-1}]] = \sigma_{s_d}^2 - \text{tr}((K_d + \sigma_n^2 I)^{-1} W)
\end{equation}

 \subsection{Surrogate loss functions for the safety objective}
 
 As discussed previously, the probability of the system violating the constraints during an episode has a dual role:
 \begin{itemize}
 \item In the policy evaluation - policy improvement iterations, it is a component of the objective function, along with the expected reward
 \item In the safety check step, where it is evaluated in order to decide whether the policy is safe enough to be  implemented in the physical system.
 \end{itemize}

According to the problem formulation, we have to keep the probability of violating the constraints under a tolerable risk threshold, every time the learning algorithm interacts with the physical system. This requirement corresponds to the second case described above. In the first case on the other hand, evaluating this probability is not a necessity as long as the policy resulting from the optimisation procedure verifies this safety requirement. Therefore, we explore substituting this probability $Q^\pi(\theta)$ with a surrogate loss function, denoted $\hat{Q}^\pi(\theta)$, investigating whether this substitution can increase computational efficiency without impeding performance. Indeed, it can potentially increase performance, if the surrogate creates such a loss landscape that facilitates the optimisation process. We can interpret the exponential penalties used in \cite{pilco2} as an example of a surrogate loss function, and we evaluate its performance in the experiments that follow.

The above considers two loss functions so far, a scaled version of the probability itself (no surrogate in this case) and exponential penalties. We tested two more: an additive cost function based on the sum of the probabilities of violating the constraints on each time step (ProbAdd for short), and one based one taking the logarithm of the original (multiplicative) probability (LogProb). 

ProbAdd is defined as: 
\begin{equation}
  \hat{Q}^\pi(\theta)_{\text{add}}  = \sum_{t=1}^T q(x_t)
\end{equation}
with $q(x_t)$ as defined in (\ref{qs})

Assuming perfect safety is feasible, with all $q(x_t) = 1$,  maximising this surrogate loss function leads to a maximisation of the original objective, which is the product of  $q(x_t)$ for all $t$. When this isn't feasible on the other hand, the maximum is not necessarily the same.

Taking the logarithm on the other hand allows us to create an additive cost function, while maintaining the same maximum, since the logarithm is a concave function. It is defined as:
\begin{equation}
  \hat{Q}^\pi(\theta)_{\text{log}}  = \log \prod_{t=1}^T q(x_t) = \sum_{t=1}^T \log (q(x_t))
\end{equation}
Differentiating the above costs is not significantly different than the process described in paragraph 4.4, and the gradients are used in the same optimisation algorithm.
 
Results are presented in Table 2. The four cost functions are evaluated on 32 runs, with 15 maximum allowed interactions per run with the physical system (480 interactions allowed in total). We can see that all methods achieve a fairly low number of collisions (way lower than the maximum allowed risk of 10\%), and the Prob and Log surrogate cost functions achieve slightly better performance. The Log and the ProbAdd cost functions fail to propose a controller for a non-negligible number of cases (4 and 6 out of 32), a fact suggesting that the optimisation proved harder using these cost functions. In the main body of the paper, the Prob cost function is used, since it achieves good performance and is in general a simpler approach, in the sense that objective (low risk) and cost function match, other than a multiplicative constant ($\xi$).

Here we examine the effects of the method we introduced for tuning the hyperparameter $\xi$. We compare the simplest version that uses a fixed value for $\xi$ (the algorithm is the same with \cite{pilco2}), with a version of the algorithm that only checks whether the policy is safe and increases $\xi$ if not (but never decreases it), and finally with the version that adapts it either by increasing or decreasing its value accordingly. To minimise other differences we use the same surrogate loss in all cases, using exponential penalties on the unsafe parts of state space.

As we see in Table 3, using an adaptively tuned $\xi$, leads to similar or better performance and more robustness to the initial choice of $\xi$.

\subsection{Experiments - details and hyperparameters}
\subsubsection{Collision Avoidance}
The original dynamical system is a very simple linear system, described by $\dot{x} = Ax + Bu$ with 
$$
A=
\begin{bmatrix}
0 & 1 & 0 & 0 \\
0 &-\frac{b}{M} & 0 & 0 \\
0 & 0 & 0 & 1 \\
0 & 0 & 0 & 0 \\
\end{bmatrix}
$$
and $B = [0, 1, 0, 0]^T$. Table \ref{table:colAvParams} includes the values for other relevant parameters.
The multiple (6) values for the initial state mean correspond to different variations
of the collision avoidance scenario. 
Performance results reported in Section \ref{sec:mainExp} are averaged over these variations.
\begin{table}
	\centering
	\resizebox{\linewidth}{!}{%
		\begin{tabular}{c c L} \hline
			\raggedleft
			Notation  & Description          & Value  \\
			b         & Friction Coefficient & 1.0    \\
			M		  & Car Mass				& 1000   \\
			J		  & \# of initial rollouts& 1      \\
			-		  & Type of Controller   & RBF    \\
			bf 		  & \# of RBF basis functions & 50 \\
			dt		  & sampling period      & 0.5(s) \\
			T 	      & episode duration	    & 25(s)  \\
			H		  & time steps per episode& 50     \\ 
			$\mu_{0,1}$ & initial state mean   & $10[-5,1,-5,1]$ \\
			$\mu_{0,2}$ & initial state mean   & $10[-5,1,-6,1]$ \\
			$\mu_{0,3}$ & initial state mean   & $10[-5,1,-7,1]$ \\
			$\mu_{0,4}$ & initial state mean   & $10[-6,1,-5,1]$ \\
			$\mu_{0,5}$ & initial state mean   & $10[-6,1,-7,1]$ \\
			$\mu_{0,6}$ & initial state mean   & $10[-7,1,-7,1]$ \\
			$\Sigma_0$  & initial state variance & $\text{diag}([1, 0.01, 1, 0.01])$ \\
			maxiter     & iterations of L-BFGS-B & 50 \\
			$\epsilon$	& max tolerable risk   & 0.10 \\
			\hline \\
	\end{tabular}}
	\captionsetup{width=0.45\textwidth}
	\caption{\small{Parameters for the collision avoidance scenario.}}
	\label{table:colAvParams}
\end{table}

\subsubsection{Building Automation Systems}
Detailed description of the simulated model used can be found in \citet{cauchi2018benchmarks}.
The simulator's parameters were trained on data from an experimental setup within the 
Department of Computer Science at the University of Oxford. 
The library provided with \cite{cauchi2018benchmarks} allows the user to create new models, with multiple rooms, 
independent or joint temperature control, deterministic dynamics, or stochastic, 
wither with additive or multiplicative disturbances.
For our experiment we use one of the predefined models, simulating two rooms, with joint temperature control,
and additive disturbances present (see section 3.2 in \cite{cauchi2018benchmarks}). 
The simulator models the dynamics as a linear time-invariant dynamical system,
described by $x[k+1] = Ax[k] + Bu[k] + Fd[k] + Q_c$, where matrices $A,B$, according to usual conventions,
refer to the inherent system dynamics with $u[k]$ being the controller's input, while $F,q$ model the 
effects of additive disturbances and $Q_c$ captures constant terms of the model.
For the exact numerical values of these parameters see the Appendix of \citet{cauchi2018benchmarks}. 

Note that in the original publication \cite{cauchi2018benchmarks} not all state variables are
observable. The observations come as $ y = Cx $, with $C=[1, 0, 0, 0, 0, 0, 0]$. 
Since our algorithm requires full observability we assume access to the full state $x$. 
This is one of the reasons that our results are not directly comparable with that of other studies in the domain 
(see \citet{abate2018arch} for an evaluation of various methods);
the second main reason is that most of these approaches assume an \emph{a priori} existing model.
With the dynamical system defined, we set the initial and target state, and the cost function as described in Section \ref{sec:BASB}.
Table \ref{table:BASBParams} includes further parameter values used for the experiments in that Section.
\begin{table}
	\centering
	\resizebox{\linewidth}{!}{%
		\begin{tabular}{c c L} \hline
			\raggedleft
			Notation & Description          & Value  \\
			J		 &\# of initial rollouts& 6      \\
			N        & \# training episodes per run & 5 \\ 
			-		 & Type of Controller   & Linear    \\
			dt		 & sampling period      & 15 (min) \\
			T 	     & episode duration	    & 12 (h)  \\
			H		 & time steps per episode& 48     \\
			$\mu_0$  & initial state mean   & \small{$[15, 20, 18,\dots,18]$} \\
			$\Sigma_0$ & initial state variance & $0.2I_d$ \\
			maxiter  & iterations of L-BFGS-B & 50 \\
			$\epsilon$ & max tolerable risk   & 0.05 \\
			\hline \\
	\end{tabular}}
	\captionsetup{width=0.45\textwidth}
	\caption{\small{Parameters for the BAS scenario.}}
	\label{table:BASBParams}
\end{table}

\subsubsection{Swimmer}

The swimmer environment we use is coming unmodified from the OpenAi-gym Python package, 
and is using the Mujoco physics simulator. 

The parameters used for training are summarised in Table \ref{table:Swimmer}.
\begin{table}
	\centering
	\resizebox{\linewidth}{!}{%
		\begin{tabular}{c c L} \hline
			\raggedleft
			Notation & Description          & Value  \\
			J		 &\# of initial rollouts& 10      \\
			N        & \# training episodes per run & 10 \\ 
			-		 & Type of Controller   & RBF    \\
			bf 		 & \# of RBF basis functions & 30 \\
			H		 & time steps per episode& 25     \\
			$\mu_0$  & initial state mean   & $\boldsymbol{0}$\\
			$\Sigma_0$ & initial state variance & $0.1I_d$ \\
			maxiter  & iterations of L-BFGS-B & 40 \\
			$\epsilon$ & max tolerable risk   & 0.10 \\
			\hline \\
	\end{tabular}}
	\captionsetup{width=0.45\textwidth}
	\caption{\small{Parameters for the Swimmer experiment.}}
	\label{table:Swimmer}
\end{table}

\FloatBarrier

\end{document}